\documentclass{Interspeech}

\interspeechcameraready 

\usepackage{array, amssymb, colortbl, pifont, booktabs, makecell, amsmath, graphicx, times, epsfig, multirow, xcolor}
\usepackage{tipa}  
\usepackage[T1]{fontenc} 
\definecolor{LightBlue}{rgb}{0.85,0.92,0.96}
\definecolor{custom_gray}{gray}{0.92}
\definecolor{darkgreen}{RGB}{0,130,0}
\definecolor{darkred}{RGB}{180,0,0}
\definecolor{cellcol}{gray}{0.92}

\title{Learning Phonetic Context-Dependent Viseme \\for Enhancing Speech-Driven 3D Facial Animation}
\author[affiliation={1}]{Hyung Kyu}{Kim}
\author[affiliation={2}]{Hak Gu}{Kim}

\affiliation{Department of Imaging Science and Arts}{Chung-Ang University}{South Korea}
\affiliation{Department of Metaverse Convergence}{Chung-Ang University}{South Korea}
\email{hyung1208@cau.ac.kr, hakgukim@cau.ac.kr}
\begin{document}
%
\maketitle

\begin{abstract}
    Speech-driven 3D facial animation aims to generate realistic facial movements synchronized with audio. Traditional methods primarily minimize reconstruction loss by aligning each frame with ground-truth. However, this frame-wise approach often fails to capture the continuity of facial motion, leading to jittery and unnatural outputs due to coarticulation. To address this, we propose a novel phonetic context-aware loss, which explicitly models the influence of phonetic context on viseme transitions. By incorporating a viseme coarticulation weight, we assign adaptive importance to facial movements based on their dynamic changes over time, ensuring smoother and perceptually consistent animations. Extensive experiments demonstrate that replacing the conventional reconstruction loss with ours improves both quantitative metrics and visual quality. It highlights the importance of explicitly modeling phonetic context-dependent visemes in synthesizing natural speech-driven 3D facial animation.
    Project page: \url{https://cau-irislab.github.io/interspeech25/}
\end{abstract}
%
%

\section{Introduction}
\label{sec:intro}
Speech-driven 3D facial animation aims to predict realistic 3D facial deformation fields, which change a given static facial mesh template, synced with input audio.
As this task is often regarded as a key generative AI technology for immersive applications such as VR remote presence, filmmaking, and game character animation \cite{application2017, application2018, kim2024analyzing, eungi2024enhancing}, it has been actively explored in recent speech and vision research.
To produce natural speech-driven 3D facial animation, a generative model is required not only to represent the 3D facial dynamics in depth beyond the vertex or mesh level, but also to understand the visible movements of vocal tract articulators during speech production \cite{facial2006}.

\begin{figure}[!t]
\begin{center}
\includegraphics[width=1.0\linewidth]{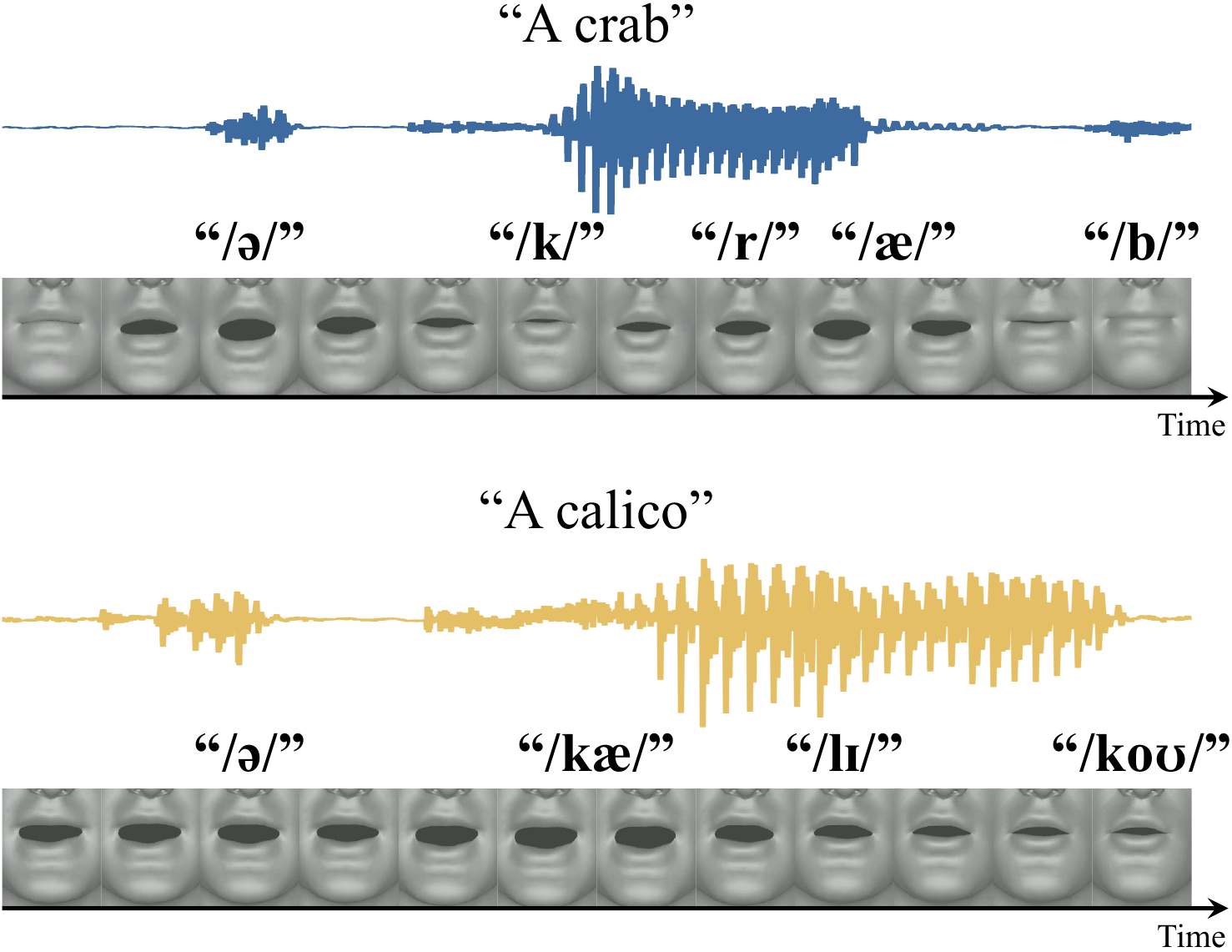}
\end{center}
\caption{Visualization of the audio and viseme when pronouncing the word "A crab" and "A calico". 
Note that the phoneme represents the corresponding dominant viseme.}
\vspace{-10pt}
\label{fig:1}
\end{figure}

We focus on addressing the \textit{coarticulation} issue for realistic speech-synchronized 3D facial animation.
Coarticulation is a phenomenon where the articulation of a speech segment is influenced by both the preceding segment (backward coarticulation) and the following segment (forward coarticulation), resulting in smoother and more natural speech transitions \cite{coarticulation24}.
As an example, Fig. \ref{fig:1} shows that the lip movements for the word ``A’’ vary according to the surrounding words spoken after it.
When pronouncing ``A crab’’, the lips tend to close after saying ``A’’.
On the other hand, the lips form a widely open shape when saying ``A calico’’.
Each viseme corresponding to a phoneme emerges gradually rather than appearing abruptly, ensuring smooth transitions between visemes.
That is, a viseme is influenced not only by the phoneme being uttered but also by its surrounding phonetic context due to the continuous lip motion and inertia of the articulators, i.e., \textit{phonetic context-dependent viseme} \cite{a2valignment17, context_dependent_viseme18}.

Previous works for speech-driven 3D facial animation used temporal-specific networks such as Temporal CNN (T-CNN), LSTM, and Transformer to capture the dynamics of audio \cite{probabilisticspeechdriven3dfacial23, CompositeandRegionalFacialMovements, facetalk2024, unitalker2024, imitator2023, diffposetalk2024}.
However, similar to motion estimation in 2D video, these methods primarily focus on reconstruction quality at each time step and facial dynamics.
As a result, they often produce jittery frames and inaccurate lip synchronization because they do not explicitly account for the various smooth viseme transitions caused by coarticulation.

With the importance of coarticulation, in this paper, we propose a simple yet effective method that explicitly considers phonetic context in speech-driven 3D facial animation.
To make the model learn the phonetic context-dependent viseme during training, we design a novel objective function for viseme coarticulation.
Unlike the conventional reconstruction loss used in existing methods, we first quantitatively measure how much the facial vertices moved within a temporal window to consider the surrounding lip movements (i.e., phonetic context-dependent viseme).
By normalizing it, we can quantify the extent of the relationship with the surrounding phonetic context on each viseme over time. We call it viseme coarticulation weight.
Then, by applying our viseme coarticulation weight to the reconstruction loss, the phonetic context-aware loss can be defined.
Extensive experiments demonstrate that learning the speech-driven facial animation models with our phonetic context-aware loss can improve their performance quantitatively and qualitatively.

Our contributions are summarized as follows:
\begin{itemize}
    \item We introduce a simple but effective objective function named as \textit{phonetic context-aware loss} to learn phonetic context-dependent viseme for addressing the coarticulation issue and enhancing speech-driven 3D facial animation.
    \item We demonstrate the effectiveness of the proposed objective function by replacing the conventional reconstruction loss in the recent models with the phonetic context-aware loss through extensive experiments on various datasets.
\end{itemize}


\begin{table*}[!th]
    \centering
    \fontsize{5pt}{7pt}\selectfont 
    \renewcommand{\arraystretch}{1.0} 
    \setlength{\tabcolsep}{4pt}
    \setlength{\fboxsep}{2pt} 
    \resizebox{0.85\textwidth}{!}{
    \begin{tabular}{l l c c c c}
    \toprule
    \multirow{2}{*}{Dataset}
     & \multirow{2}{*}{Method} 
     & \multicolumn{4}{c}{Objective Function (Original / \colorbox{custom_gray}{Ours})} \\
    \cmidrule(lr){3-6}
     &  & FVE $\downarrow$ & LVE $\downarrow$ & LDTW $\downarrow$ & Lip-max $\downarrow$ \\
    \midrule
    \multirow{4}{*}{VOCASET \cite{VOCA2019}}
      & FaceFormer \cite{faceformer2022}
        & 0.637 / \colorbox{custom_gray}{\textbf{0.633}}
        & 0.414 / \colorbox{custom_gray}{\textbf{0.397}}
        & 0.482 / \colorbox{custom_gray}{\textbf{0.465}}
        & 0.618 / \colorbox{custom_gray}{\textbf{0.608}} \\
      & CodeTalker \cite{codetalker2023}
        & 0.690 / \colorbox{custom_gray}{\textbf{0.642}}
        & 0.468 / \colorbox{custom_gray}{\textbf{0.425}}
        & 0.496 / \colorbox{custom_gray}{\textbf{0.464}}
        & 0.631 / \colorbox{custom_gray}{\textbf{0.615}} \\
      & SelfTalk \cite{selftalk23}
        & 0.606 / \colorbox{custom_gray}{\textbf{0.590}}
        & 0.395 / \colorbox{custom_gray}{\textbf{0.368}}
        & 0.460 / \colorbox{custom_gray}{\textbf{0.444}}
        & 0.591 / \colorbox{custom_gray}{\textbf{0.587}} \\
      & ScanTalk \cite{scantalk2024}
        & 0.624 / \colorbox{custom_gray}{\textbf{0.590}}
        & 0.428 / \colorbox{custom_gray}{\textbf{0.359}}
        & 0.507 / \colorbox{custom_gray}{\textbf{0.460}}
        & 0.600 / \colorbox{custom_gray}{\textbf{0.570}} \\
    \midrule
    \multirow{3}{*}{BIWI \cite{BIWI2010}}
      & FaceFormer \cite{faceformer2022}
        & 0.992             / \colorbox{custom_gray}{\textbf{0.964}}
        & 0.212             / \colorbox{custom_gray}{\textbf{0.209}}
        & \textbf{0.140}    / \colorbox{custom_gray}{\textbf{0.140}}
        & 0.375             / \colorbox{custom_gray}{\textbf{0.373}}  \\
      & CodeTalker \cite{codetalker2023}
        & 0.979 / \colorbox{custom_gray}{\textbf{0.929}}
        & 0.207 / \colorbox{custom_gray}{\textbf{0.195}}
        & 0.143 / \colorbox{custom_gray}{\textbf{0.139}}
        & 0.377 / \colorbox{custom_gray}{\textbf{0.371}}  \\
      & SelfTalk \cite{selftalk23}   
        & 1.110 / \colorbox{custom_gray}{\textbf{0.933}}
        & 0.240 / \colorbox{custom_gray}{\textbf{0.201}}
        & 0.150 / \colorbox{custom_gray}{\textbf{0.137}}
        & 0.401 / \colorbox{custom_gray}{\textbf{0.364}}  \\
    \midrule
    \multirow{1}{*}{$\text{BIWI}_6$ \cite{scantalk2024, BIWI2010}}
      & ScanTalk \cite{scantalk2024}
        & 0.463 / \colorbox{custom_gray}{\textbf{0.454}}
        & 0.114 / \colorbox{custom_gray}{\textbf{0.112}}
        & 0.112 / \colorbox{custom_gray}{\textbf{0.110}}
        & 0.837 / \colorbox{custom_gray}{\textbf{0.827}}  \\
    \midrule
    \multirow{1}{*}{Multiface \cite{multiface2022}}
      & ScanTalk \cite{scantalk2024}
        & 0.551 / \colorbox{custom_gray}{\textbf{0.531}}
        & 0.100 / \colorbox{custom_gray}{\textbf{0.092}}
        & 0.104 / \colorbox{custom_gray}{\textbf{0.099}}
        & 0.821 / \colorbox{custom_gray}{\textbf{0.802}}  \\
    \bottomrule
    \end{tabular}
    }
    \caption
    {
    Quantitative evaluations for 3D facial animations before and after replacing the original reconstruction loss with our phonetic context-aware loss on existing models (Original / \colorbox{custom_gray}{Ours}). Lower values indicate better performance.
    }
    \vspace{-10pt}
    \label{table:1}
\end{table*}

\section{Method}
In this section, we illustrate the proposed method with the common objective functions used in existing speech-driven 3D facial animator baselines such as FaceFormer \cite{faceformer2022} and CodeTalker \cite{codetalker2023}.
This section consists of two subsections: 1) common objective functions in speech-driven 3D facial animator baselines as a preliminary and 2) learning phonetic context-dependent viseme to improve the baselines with the proposed phonetic context-aware loss.

\subsection{Objective Functions in Speech-Driven 3D Facial Animator Baselines}
Most existing speech-driven 3D facial animation models mainly focus on minimizing differences between the ground-truth and the predicted 3D facial vertices equally at each time step.
To strictly align with the ground-truth for each time $t$, they employ the reconstruction loss $\mathcal{L}_{rec}$, which is the Euclidean distance between ground-truth and the generated 3D facial vertices.
\begin{equation}
    \mathcal{L}_{rec} = \sum_{t=1}^T \bigl\lVert v^t - \hat{v}^t \bigr\rVert^2,
\end{equation}
where $v^t$ and $\hat{v}^t$ are ground-truth and the synthesized 3D facial vertices at time $t$, respectively. $T$ is the length of the sequence.
Note that $\hat{v}^t$ is obtained by adding the predicted 3D facial deformation fields $\hat{d}^t$ to a given static 3D facial template as in speech-driven 3D facial animator baselines \cite{faceformer2022,codetalker2023,selftalk23,scantalk2024}.

Utilizing the reconstruction loss $\mathcal{L}_{rec}$ alone for optimization can induce jittery output because it does not take into account facial dynamics in the time domain.
In addition, it uniformly processes the local dynamics at each time $t$, meaning that it disregards the phonetic context.

To mitigate the problem of jittery output frames caused by relying solely on reconstruction loss $\mathcal{L}_{rec}$ \cite{selftalk23}, the velocity loss $\mathcal{L}_{vel}$ is generally employed to promote smoother and more natural lip movements over time.
The velocity loss can be written as
\begin{equation}
    \mathcal{L}_{vel} = \sum_{t=2}^{T} \bigl\lVert (v^t - v^{t-1}) - (\hat{v}^{t} - \hat{v}^{t-1}) \bigr\rVert^2.
\end{equation}

\subsection{Learning Phonetic Context-Dependent Viseme}
The recent speech-driven 3D facial animators have demonstrated remarkable performance.
However, it is insufficient to merely minimize the Euclidean distance between the ground truth and predicted facial vertices, as well as the dynamics of consecutive frames over time for producing clear and intelligible lip movements. To synthesize a realistic and natural 3D facial animation, we introduce a novel phonetic context-aware loss that attends to variations of facial movements in both the preceding speech segment (backward coarticulation) and the following speech segment (forward coarticulation).
In various audio-visual tasks, such as lip reading \cite{a2valignment17, lipreading20} and 2D video-based speech recognition \cite{a2valignment22, a2valignment23}, explicitly modeling context-dependent visemes has been shown to improve the robustness and accuracy of alignment between visual and auditory streams.

Based on these studies, to learn phonetic context-dependent viseme in a 3D facial animation, we design a novel phonetic context-aware loss $\mathcal{L}_{pc}$ that captures the characteristics of phonetic context-dependent visemes, where the articulation of a phone gradually varies depending on the preceding or following phones.
To observe surrounding phones, we first define a temporal window size, which can be written as
\begin{equation}
    \Omega_{\sigma}^{t} = \{k \text{ }|\text{ } t-\sigma \le k \le t+\sigma\ \}, \text{ } t = 1+\sigma, \cdots, T-\sigma,
\end{equation}
where $t$ is the current frame index and $T$ is the total number of frames.
$\sigma$ is the window radius that determines the temporal neighborhood around frame $t$ (i.e., window size can be obtained by $2\sigma+1$).
In this work, the window size is set to 5 ($\sigma=2$).

Then, we can measure the changes of the articulation of a phone at time $t$ within the temporal window $|\Omega_{\sigma}^{t}|$ (i.e., small speech segment).
\begin{equation}
    w^{t} = \frac{1}{|\Omega_{\sigma}^{t}|}\sum_{k\in \Omega_{\sigma}^{t}}\big|| v^k-v^{k-1}|\big|^2.
\end{equation}

The measured $w^{t}$ represents the extent of change in facial vertices within the vicinity of frame $t$ due to coarticulation.
By measuring the local facial and lip movement $w^t$ around time $t$ and normalizing it over time, we can approximately estimate the extent of the phonetic context-dependent viseme.

With the normalization, finally, the proposed viseme coarticulation weight $\tilde{w}^{t}$ can be defined as
\begin{equation}
    \label{eq:normalized-weight}
    \tilde{w}^t = \frac{\exp\bigl(w^t\bigr)}{\sum_{i=1}^{T} \exp\bigl(w^{i}\bigr)}.
\end{equation}

The proposed viseme coarticulation weight $\tilde{w}^{t}$ focuses more on regions where articulatory movement changes relatively more due to coarticulation. By applying the weight $\tilde{w}^{t}$ to the conventional reconstruction loss $\mathcal{L}_{rec}$, we can define the proposed phonetic context-aware loss $\mathcal{L}_{pc}$. 

\begin{equation}
    \label{eq:viseme-coarticulation-loss}
    \mathcal{L}_{pc}
    = \sum_{t=1}^{T} \tilde{w}^t \cdot \bigl\lVert v^t - \hat{v}^t \bigr\rVert^2,
\end{equation}

In training stage, by minimizing $\mathcal{L}_{pc}$, the model can generate more natural 3D facial animations with continuous and smooth transition.

\section{Experiments}
\label{sec:Experiments}
\begin{figure*}[t!]
\centering
\includegraphics[width=0.9\linewidth]{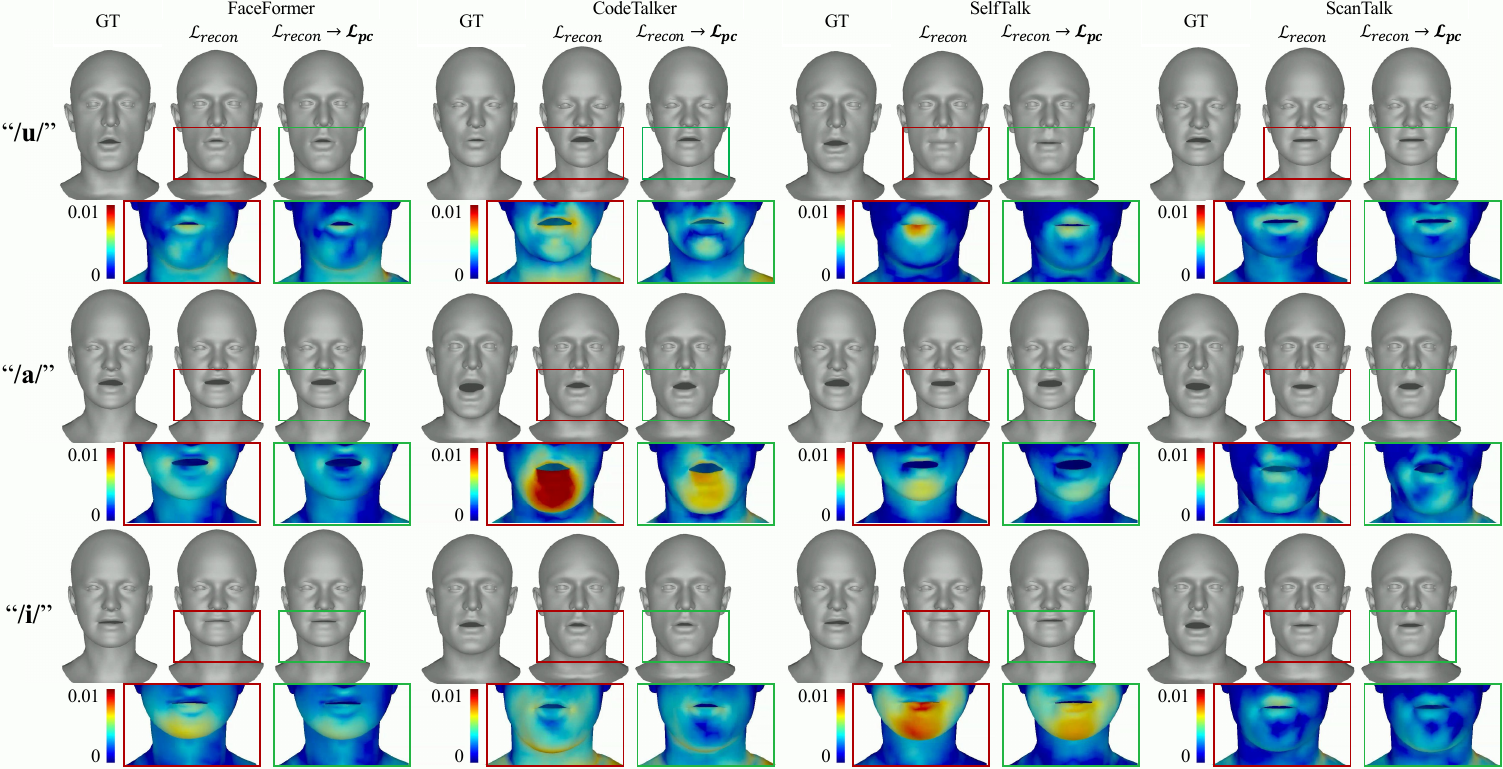}
\caption
{
Visual comparisons of baseline models vs. baseline models trained with our objective function on VOCASET. Note that the second, fourth, and sixth rows represent the visualization of per-vertex errors in lip regions.
}
\label{fig:2}
\vspace{-10pt}
\end{figure*}

\subsection{Experimental Settings}
\subsubsection{Datasets}
In our experiments, we conduct extensive experiments on four widely-used datasets, which are VOCASET \cite{VOCA2019}, BIWI \cite{BIWI2010}, $\text{BIWI}_6$ \cite{scantalk2024, BIWI2010}, and MultiFace \cite{multiface2022}.
These datasets include pairs of audio and the corresponding 3D facial scans that show the pronunciation of English speech.

\noindent \textbf{VOCASET.} VOCASET comprises a total of 480 facial motion sequences from 12 people, recorded at 60 frames per second for roughly 4 seconds each, and 255 distinct words, 5 sentences of which are shared by all speakers. Every 3D face model has 5,023 vertices and is registered to the FLAME \cite{FLAME2017} topology.

\noindent \textbf{BIWI.} BIWI consists of 40 unique sentences shared across all speakers.
There are 40 sentences uttered by 14 subjects. 
A dynamic 3D facial scan at 25 frames per second is captured by repeating each recording twice in either an emotional or neutral setting. 
There are 23,370 vertices in the registered topology, and the average sequence length is approximately 4.67 seconds. 

\noindent $\textbf{BIWI}_\textbf{6.}$ $\text{BIWI}_6$ is a down-sampled version of BIWI with a fixed topology and 3,895 vertices and 7,539 faces used in \cite{scantalk2024}.

\noindent \textbf{Multiface.} Multiface includes 13 persons delivering up to 50 utterances, each around 4 seconds in duration, sampled at 30 frames per second and possess a static topology including 5,471 vertices and 10,837 faces.

\subsubsection{Models}
\noindent To verify the effectiveness of the proposed phonetic context-aware loss, we employ four recent speech-driven 3D facial animators as our baseline models. When training the model with our objective function, the conventional reconstruction loss is replaced with our phonetic context-aware loss.

\noindent \textbf{FaceFormer.} FaceFormer \cite{faceformer2022} introduces a transformer-based 3D facial animator to capture the relationship between audio and previous motions to generate 3D facial animations.

\noindent \textbf{CodeTalker.} CodeTalker \cite{codetalker2023} discretely stores 3D motion priors using the VQ-VAE structure, which is known to be effective for image restoration.

\noindent \textbf{SelfTalk.} SelfTalk \cite{selftalk23} can restore more realistic lip shapes by learning to make the lip-reading result distribution of the generated mesh similar to the ASR result distribution of the audio using the commutative diagram structure.

\noindent \textbf{ScanTalk} \cite{scantalk2024} introduces a diffusion-based approach to enable facial animation synthesis for various topologies with a single model regardless of the topology between various datasets.

\subsubsection{Metrics}
For quantitative evaluation, we employ four metrics, which are FVE, LVE, LDTW, and Lip-max.

\noindent \textbf{Face Vertex Error (FVE).} FVE quantifies the geometric discrepancy between a reference and the generated meshes by Euclidean distance for each vertex across the entire facial region.

\noindent \textbf{Lip Vertex Error (LVE).} LVE quantifies the Euclidean distance for each vertex specifically within the lip region.

\noindent \textbf{Lip Dynamic Time Warping (LDTW).} LDTW is an index that measures the similarity of lip movements over time using dynamic time warping (DTW) \cite{dtw07, imitator2023} for temporal consistency.

\noindent \textbf{Lip-max.} Lip-max is a metric used in \cite{meshtalk2021, imitator2023} that calculates the largest vertex error within the lip region.

\begin{figure}[t!]
\centering
\includegraphics[width=1.0\linewidth]{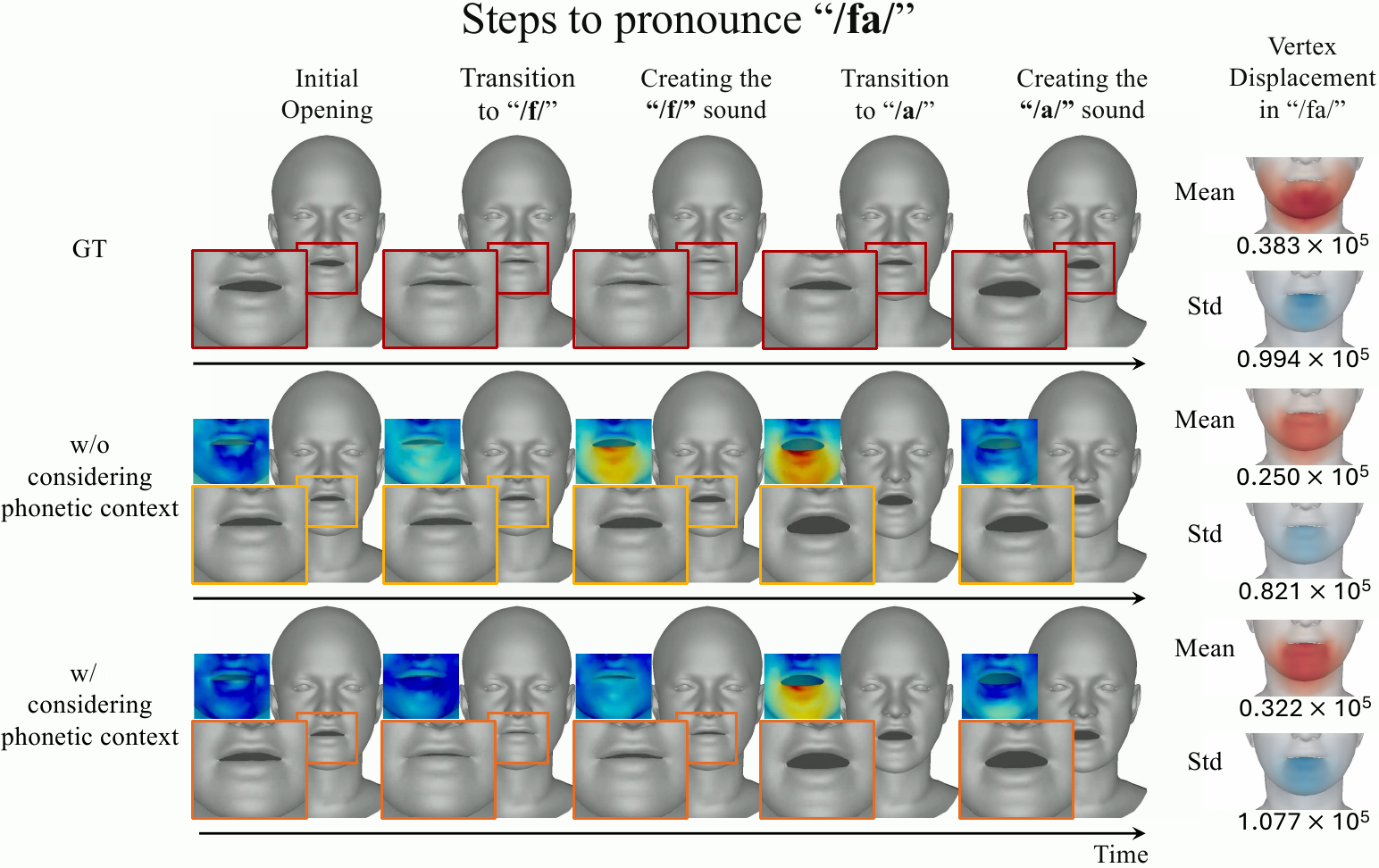}
\caption
{
Qualitative comparison of optimizing the baseline models with their original objective functions vs. with our phonetic context-aware loss instead of reconstruction loss.
}
\label{fig:3}
\vspace{-10pt}
\end{figure}

\subsection{Quantitative Results}

For performance evaluations of the proposed phonetic context-aware loss, we trained the baseline models with $\mathcal{L}_{pc}$ instead of $\mathcal{L}_{rec}$. Except for the reconstruction loss $\mathcal{L}_{rec}$, we did not change other objective functions used in the official code of each baseline model.
To quantitatively evaluate the performance of proposed method, we calculate FVE, LVE, LDTW, and Lip-max as seen in Tab. \ref{table:1}.
In our experiments, we follow the official split criteria used in each baseline model for training and inference.
As seen in Tab. \ref{table:1}, the models trained with our phonetic context-aware loss achieved better performance, compared to those of their original versions without any modifications.
This means that, instead of treating the error of each facial vertex equally, considering the importance of changes in visemes based on phonetic context enables the creation of more natural speech-driven 3D facial animations.

\subsection{Qualitative Results}

Fig. \ref{fig:2} shows visual comparisons between the ground-truth, FaceFormer \cite{faceformer2022} w/ $\mathcal{L}_{rec}$ or $\mathcal{L}_{pc}$, CodeTalker \cite{codetalker2023} w/ $\mathcal{L}_{rec}$ or $\mathcal{L}_{pc}$, SelfTalk \cite{selftalk23} w/ $\mathcal{L}_{rec}$ or $\mathcal{L}_{pc}$, and ScanTalk \cite{scantalk2024} w/ $\mathcal{L}_{rec}$ or $\mathcal{L}_{pc}$. 
It is essential to consider phonetic context-dependent viseme and as a results, it enables achieving higher vertex alignment.

Fig. \ref{fig:3} shows an example of the pronunciation of ``/fa''. In Fig. \ref{fig:3}, the sixth column visualizes the mean and variance of the movement of vertices at the time of the pronunciation. When considering phonetic context-dependent viseme, more accurate viseme can be generated in the transition region from one phoneme to another than when not considering context. 
Importantly, learning phonetic context-dependent viseme can produce more perceptually natural and smooth transition, compared to the existing models that did not account for coarticulation.

\begin{figure}[!t]
\begin{center}
\includegraphics[width=1.0\linewidth]{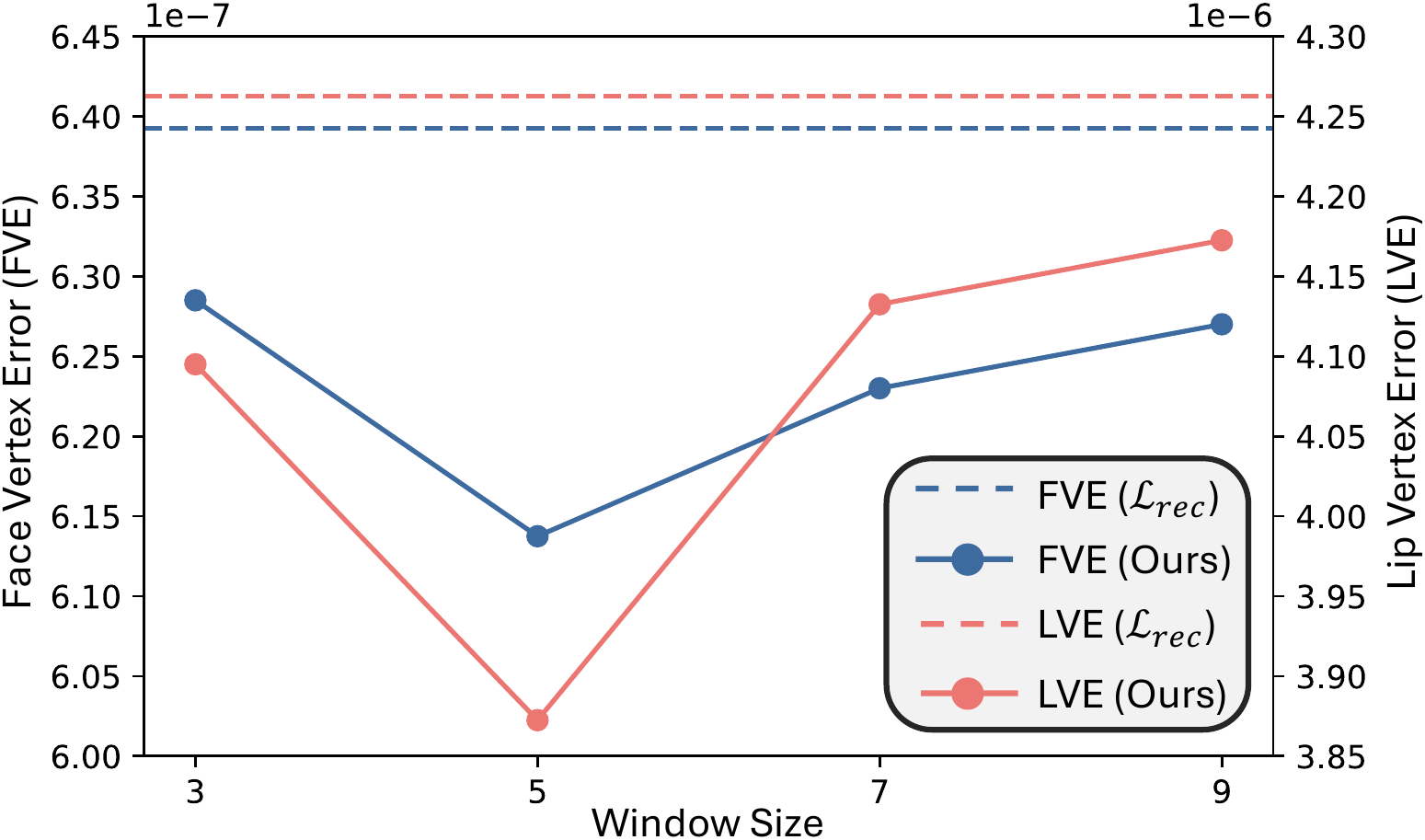}
\end{center}
\caption
{
Performance analysis according to different window sizes of the viseme coarticulation weight in \textcolor[RGB]{72, 106, 155}{FVE} and \textcolor[RGB]{221, 124, 118}{LVE}.
}
\label{fig:4}
\vspace{-10pt}
\end{figure}

\subsection{Ablation Study}
We conduct performance analysis to demonstrate the impact of the window size in the proposed viseme coarticulation weight.
Specifically, we investigate both FVE and LVE on VOCASET \cite{VOCA2019} when we train baselines with or without the proposed phonetic context-aware loss according to different window sizes.
In Fig. \ref{fig:4}, the dotted line represents the average performance of original baseline models trained using their official code. In contrast, the solid line represents the average performance of the same baseline models trained with our objective function instead of the reconstruction loss.
As shown in Fig. \ref{fig:4}, when considering the phonetic context in synthesizing facial animation, both FVE and LVE values are much lower than when it is ignored, regardless of window sizes (i.e., the length of small speech segment for phonetic context). In our experiments, we set the window size to 5, leading to the lowest FVE and LVE.

\section{Conclusion}
\label{sec:Conclusion}
In this paper, we introduced a phonetic context-aware loss to enhance the naturalness of speech-driven 3D facial animation by explicitly modeling viseme transitions influenced by phonetic context. Unlike existing approaches that treat all facial vertices equally in the reconstruction loss, our method leverages a viseme coarticulation weight, which assigns greater importance to regions with pronounced articulatory changes. Experimental results across multiple datasets confirm that our approach not only reduces prediction errors but also produces smoother and more realistic facial animations. Additionally, our ablation study shows that the window size for capturing phonetic context plays a crucial role in optimizing animation quality. These findings emphasize the necessity of incorporating phonetic context modeling in speech-driven 3D animation frameworks. Future work will explore extending our approach to more diverse linguistic contexts and integrating additional multimodal cues for further improvements.

\section{Acknowledgments}
This research was supported by the National Research Foundation of Korea(NRF) grant funded by the Korea government(MSIT) (RS-2023-00253232) and Culture, Sports and Tourism R\&D Program through the Korea Creative Content Agency grant funded by the Ministry of Culture, Sports and Tourism in 2023 (Project Name: Acquisition of 3D precise information of microstructure and development of authoring technology for ultra-high prediction cultural restoration, Project Number: RS-2023-00227749).


\label{sec:refs}

\bibliographystyle{IEEEbib}
\bibliography{ref}

\end{document}